\documentclass[letterpaper]{article} 
\usepackage{aaai2026}  
\usepackage{times}  
\usepackage{helvet}  
\usepackage{courier}  
\usepackage[hyphens]{url}  
\usepackage{graphicx} 
\usepackage{natbib}  
\usepackage{caption} 
\usepackage{algorithm}
\usepackage{algorithmic}

\usepackage{newfloat}
\usepackage{listings}
\DeclareCaptionStyle{ruled}{labelfont=normalfont,labelsep=colon,strut=off} 
\floatstyle{ruled}
\newfloat{listing}{tb}{lst}{}
\floatname{listing}{Listing}
\usepackage{amsmath}
\usepackage{amssymb}
\usepackage{booktabs}
\usepackage{arydshln}
\usepackage{multirow}

\title{Enhancing Meme Emotion Understanding with Multi-Level\\Modality Enhancement and Dual-Stage Modal Fusion}
\author{
    Yi Shi\textsuperscript{\rm 1},
    Wenlong Meng\textsuperscript{\rm 1},
    Zhenyuan Guo\textsuperscript{\rm 1},
    Chengkun Wei\textsuperscript{\rm 1}\thanks{Corresponding author},
    Wenzhi Chen\textsuperscript{\rm 1}
}
\affiliations{

    \textsuperscript{\rm 1}Zhejiang University\\
    \{shiyi666, mengwl, zhenyuanguo, weichengkun, chenwz\}@zju.edu.cn
}

\usepackage{bibentry}

\begin{document}

\maketitle

\begin{abstract}
With the rapid rise of social media and Internet culture, memes have become a popular medium for expressing emotional tendencies. This has sparked growing interest in Meme Emotion Understanding (MEU), which aims to classify the emotional intent behind memes by leveraging their multimodal contents. While existing efforts have achieved promising results, two major challenges remain: (1) a lack of fine-grained multimodal fusion strategies, and (2) insufficient mining of memes’ implicit meanings and background knowledge. To address these challenges, we propose \textit{MemoDetector}, a novel framework for advancing MEU. First, we introduce a four-step textual enhancement module that utilizes the rich knowledge and reasoning capabilities of Multimodal Large Language Models (MLLMs) to progressively infer and extract implicit and contextual insights from memes. These enhanced texts significantly enrich the original meme contents and provide valuable guidance for downstream classification. Next, we design a dual-stage modal fusion strategy: the first stage performs shallow fusion on raw meme image and text, while the second stage deeply integrates the enhanced visual and textual features. This hierarchical fusion enables the model to better capture nuanced cross-modal emotional cues. Experiments on two datasets, MET-MEME and MOOD, demonstrate that our method consistently outperforms state-of-the-art baselines. Specifically, \textit{MemoDetector} improves F1 scores by 4.3\% on MET-MEME and 3.4\% on MOOD. Further ablation studies and in-depth analyses validate the effectiveness and robustness of our approach, highlighting its strong potential for advancing MEU. Our code is available at https://github.com/singing-cat/MemoDetector.
\end{abstract}


\section{Introduction}

In recent years, the rise of Internet culture has led to an increasing use of memes as a medium for expressing personal emotions and intentions \cite{shifman2013memes,vasquez2021cats}. Memes, characterized by their inherently multimodal nature combining image and text, have become a central form of online communication. As a result, Meme Emotion Understanding (MEU) has emerged as an important and timely research topic. However, unlike conventional multimodal sentiment analysis, memes often rely on metaphorical expressions and are highly context-dependent, which poses a crucial challenge for MEU.

Several studies have made promising strides in meme emotion understanding. \citet{xu2022met} constructed a bilingual dataset focused on metaphor-rich memes, where emotion detection was introduced as a subtask. They also proposed a single-stage multimodal fusion method tailored to this task. Similarly, \citet{sharma2024emotion} created a dataset based on Ekman's six basic emotions \cite{ekman2011meant}, specifically designed for meme emotion classification. While these works represent meaningful progress, they still suffer from two major limitations: (1) \textbf{a lack of fine-grained multimodal fusion strategies} that can fully capture the intricate interplay between visual and textual elements in memes and (2) \textbf{insufficient mining of implicit meanings and background knowledge}, which are crucial for understanding the nuanced emotions embedded in memes.

\begin{figure}[!t]
    \centering
    \includegraphics[width=0.45\textwidth]{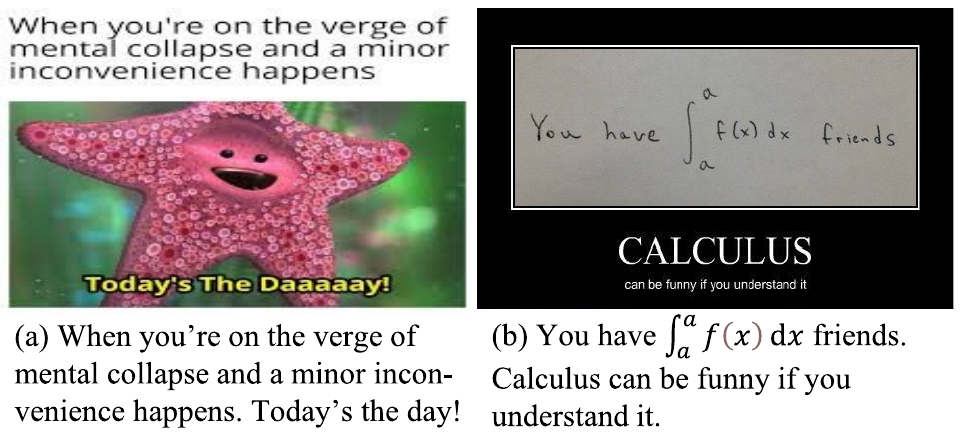}
    \caption{Examples illustrating the challenges faced by existing methods.}
    \label{fig:exp}
    \vspace{-0.4cm}
\end{figure}

On one hand, most existing approaches perform only a shallow fusion of the visual and textual modalities in memes, overlooking the subtle and intricate relationships between image and text. Such simplistic fusion strategies often limit the model's ability to capture cross-modal nuances, leading to misinterpretations of the meme's intended emotional tone. As illustrated in Figure \ref{fig:exp} (a), the meme combines a cheerful facial expression with sorrowful text, creating a contrast that effectively highlights the sender’s underlying sadness. Simple fusion strategies fail to capture this nuanced interplay between modalities, making it difficult for the model to correctly interpret the emotional message. On the other hand, existing methods focus solely on the meme itself, lacking deeper exploration of its implicit meanings and relevant background knowledge. As a result, crucial emotional cues embedded in cultural or contextual references may be overlooked. For example, as illustrated in Figure \ref{fig:exp} (b), the sentiment of this meme is embedded in the mathematical result of an integral expression. Without sufficient mathematical knowledge, small models are unlikely to recognize the underlying sadness behind this meme.

To address these challenges, we introduce \textit{MemoDetector}, a novel framework designed to advance MEU. First, we leverage the rich world knowledge and powerful reasoning capabilities of Multimodal Large Language Models (MLLMs) to perform a four-step textual enhancement of memes. These four steps progress from surface to depth, covering image, text, their joint semantics, and the meme’s potential real-world context. Through this multi-level analysis, our approach effectively uncovers the underlying meanings and background knowledge embedded in memes, significantly enriching the original information and providing strong guidance for downstream classification by a small model. Next, to address the limitations of prior approaches that rely on overly simplistic fusion strategies, we propose a dual-stage modal fusion mechanism. This strategy not only performs a shallow fusion of surface-level features but also enables a deeper, bidirectional integration of the enhanced visual and textual representations. Specifically, we first conduct a shallow fusion between original meme image and its accompanying text to obtain an enriched visual representation. We then perform a deep bidirectional fusion between the enriched visual features and the enhanced textual features, allowing both modalities to mutually refine and reinforce each other. This dual-stage fusion empowers the model to develop a more nuanced understanding of inter-modal relationships and capture subtle emotional cues that may be overlooked by conventional single-stage fusion methods.

To validate the effectiveness of our method, we conduct extensive experiments on two benchmark datasets: MET-MEME \cite{xu2022met} and MOOD \cite{sharma2024emotion}. The results demonstrate that our model significantly outperforms state-of-the-art (SoTA) baselines across all evaluation metrics on both datasets. Specifically, on MET-MEME, our approach achieves improvements of 4.17\% in Accuracy and 4.3\% in Macro-F1. On MOOD, the gains are 4.04\% and 3.4\%, respectively. In addition, we conduct ablation and in-depth analytical experiments to validate the unique contributions of each component in our framework and its superiority over existing methods. These findings further demonstrate the effectiveness and robustness of our approach.

Our main contributions can be summarized as follows:
\begin{itemize}
    \item We analyze two core challenges in meme emotion understanding, and propose a novel framework, \textit{MemoDetector}, which for the first time incorporates contextual background knowledge and introduces a fine-grained modal fusion strategy to enhance MEU.
    \item We design a four-step textual enhancement module and a dual-stage modal fusion mechanism. These components hierarchically uncover meme background knowledge and implicit meanings, while enabling a more nuanced fusion of visual and textual features.
    \item Extensive experiments demonstrate the superiority of our method, achieving state-of-the-art performance across all evaluation metrics in meme emotion understanding.
\end{itemize}

\begin{figure*}[t]
\centering
\includegraphics[width=0.9\textwidth]{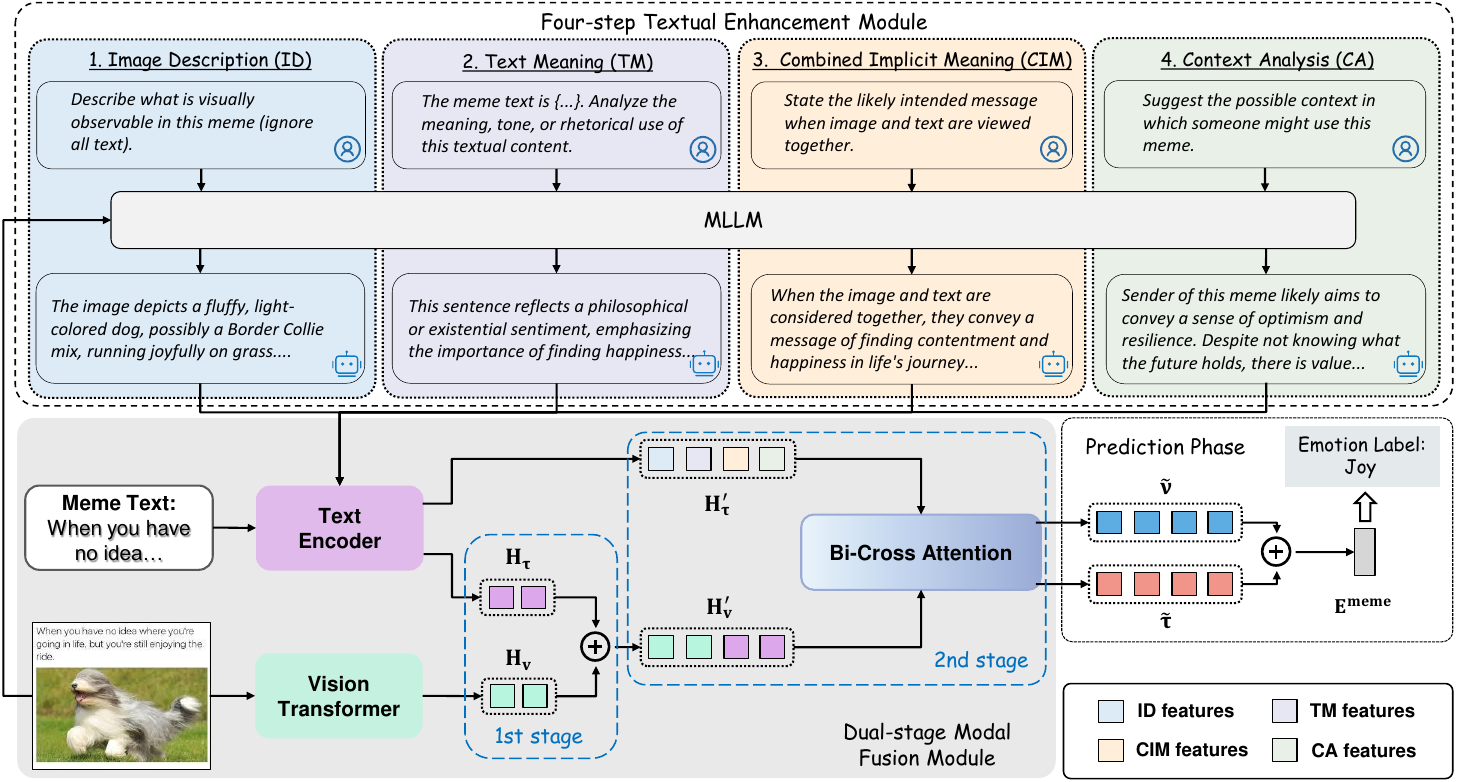} 
\caption{Overview of our framework. We first prompt MLLM to generate multi-level insights for the meme, which are distilled into a small model as auxiliary guidance. The small model then performs dual-stage modal fusion to produce the final prediction.}
\label{main}
\vspace{-4mm}
\end{figure*}

\section{Related Work}

\subsection{Multimodal Emotion Detection}

Previous studies have primarily focused on the broader task of multimodal emotion analysis \cite{hu2022unimse,ahmed2023systematic,cheng2024emotion}, rather than specifically targeting memes. Unlike unimodal emotion analysis, multimodal emotion detection requires effective fusion of features from different modalities \cite{zhu2023multimodal,gan2024multimodal}. Traditionally, multimodal fusion strategies are categorized into three types: data-level fusion \cite{danapal2020sensor}, early fusion \cite{pranesh2020memesem}, and late fusion \cite{pramanick2021detecting}. Data-level fusion merges raw inputs at the input stage, while early fusion integrates modality-specific features. In contrast, late fusion combines separate predictions via decision-level strategies. Although these fusion methods have shown effectiveness in general multimodal emotion analysis, they often fall short when applied to memes, a unique medium that frequently involves metaphorical and context-dependent cues \cite{younes2022metaphors}. These simplistic fusion methods lack the capacity to capture the intricate and implicit interplay between meme's modalities.

\subsection{Meme Analysis}

With the rise of memes, research on meme analysis has increasingly emerged. Existing studies mainly focus on hateful memes. \citet{cao2023pro} used a vision-language model to generate enriched meme captions, which were then fed into a language model for hate classification.. \citet{lin2024towards} proposed an innovative approach by leveraging LLMs to conduct debates. Beyond hate detection, finer-grained tasks like meme emotion detection have also been explored. \citet{xu2022met} introduced a meme understanding dataset that includes emotion analysis as a subtask. \citet{sharma2024emotion} proposed MOOD dataset, specifically designed for meme emotion classification. Methods for this task include both unimodal and multimodal approaches. For unimodal feature extraction, models such as BERT \cite{devlin2019bert} and GloVe \cite{pennington2014glove} were used for text, while EfficientNet \cite{tan2019efficientnet} and ResNet \cite{he2016deep} served as image encoders. Some studies, such as \citet{singh2020lt3} and \citet{vlad2020upb}, applied multi-task learning to jointly predict emotion and sentiment. Despite encouraging results, existing methods struggle to capture memes’ nuanced semantics and contextual variability, as small models lack the capacity for deep understanding.

\section{Methodology}

\subsection{Task Definition}

This study addresses the task of meme emotion understanding (MEU), formulated as a multi-class classification problem. Specifically, given a meme instance $M = (I, T)$, where $I$ denotes the image component and $T$ represents the associated textual content, the objective is to predict the corresponding emotion label $y_{emo}$ that reflects the affective intent likely conveyed by the meme's sender.

Our core idea is to harness the rich background knowledge and powerful reasoning capabilities of MLLMs \cite{liang2024survey,wang2024exploring} to perform a multi-level analysis of memes. This includes shallow perception, deep interpretation, and associative reasoning. The resulting explanations are distilled as auxiliary knowledge to a lightweight classifier, enhancing its capacity for MEU. Subsequently, the classifier employs a carefully designed dual-stage modal fusion strategy to effectively integrate information from both modalities, leading to more accurate predictions. The overview of our framework is shown in Figure \ref{main}.

\subsection{Four-step Text Modality Enhancement}

Leveraging the rich knowledge and reasoning abilities of MLLMs, we generate multi-level interpretations of memes to support more accurate emotion prediction. Traditional small-model-based approaches often capture only surface-level features of memes, lacking the ability to understand deeper aspects such as metaphor and context. Our method addresses this by providing enriched semantic insights that enhance the model's emotional understanding.

Inspired by the progressive nature of human cognition \cite{bloom1956taxonomy}, we divide MLLM's meme understanding process into three hierarchical levels: shallow perception, deep interpretation, and associative reasoning. This tiered structure enables MLLM to gradually build a more nuanced understanding. To further enhance the quality of MLLM-generated textual explanations, we adopt a Chain-of-Thought (CoT) prompting strategy \cite{wei2022chain}, designing a four-step prompt sequence aligned with the three levels of understanding: \textbf{Image Description (ID)}, \textbf{Text Meaning (TM)}, \textbf{Combined Implicit Meaning (CIM)}, and \textbf{Context Analysis (CA)}. We provide a detailed explanation of each step as follows.

\textbf{Step 1: Image Description.} To provide MLLM with a holistic understanding of memes' visual content, we begin by prompting it to describe key visual elements, such as objects, characters, and background scenes. While these visual cues represent surface-level features, they are essential for emotion recognition. For instance, facial expressions often convey strong emotional signals. To eliminate interference from meme text, we design the following prompt $p^{ID}$:

\textit{``Describe what is visually observable in this meme (ignore all text)."}

Given meme image $I$ and prompt $p^{ID}$, MLLM will generate image description $T^{ID}$ as follows:
\begin{equation}
    T^{ID} = \mathrm{MLLM}(I, p^{ID})
\end{equation}

\textbf{Step 2: Text Meaning.} In addition to visual information, textual modality also plays a crucial role in meme emotion detection. Many memes convey specific emotional intentions through cleverly crafted language and metaphorical expressions. However, meme texts often employ rhetoric or emphatic tones that obscure their underlying emotional cues. To uncover these signals, we instruct MLLM to focus on the semantics, tone, and rhetorical usage of meme text by providing the following prompt $p^{TM}$:

\textit{``The meme text is $\{...\}$. Analyze the meaning, tone, or rhetorical use of this textual content."}

Given meme text $T$ and prompt $p^{TM}$, MLLM will generate text meaning $T^{TM}$ as follows:
\begin{equation}
    T^{TM} = \mathrm{MLLM}(T, p^{TM})
\end{equation}

\textbf{Step 3: Combined Implicit Meaning.} After separately analyzing each modality, we prompt the MLLM to interpret the combined metaphorical meaning that emerges from the interplay between image and text. This step is crucial, as many memes rely on subtle cross-modal associations to convey nuanced messages that cannot be captured by unimodal analysis alone. Ignoring such interactions can result in a biased or incomplete understanding. To elicit this joint interpretation, we design the following prompt $p^{CIM}$:

\textit{``State the likely intended message when image and text are viewed together."}

Given meme image $I$, meme text $T$ and prompt $p^{CIM}$, MLLM will generate combined implicit meaning $T^{CIM}$:
\begin{equation}
    T^{CIM} = \mathrm{MLLM}(I,T, p^{CIM})
\end{equation}

\textbf{Step 4: Context Analysis.} In the final step, after MLLM has formed a relatively comprehensive understanding of the meme, we simulate human cognitive processes by encouraging associative reasoning, prompting the model to infer possible usage contexts based on its prior analysis. Given the context-dependent nature of memes, their emotional connotations often become more salient when grounded in a plausible situational backdrop. To enable the model to capture this contextual nuance, we introduce prompt $p^{CA}$:

\textit{``Suggest the possible context in which someone might use this meme."}

Given meme image $I$, meme text $T$ and prompt $p^{CA}$, MLLM will generate context analysis $T^{CA}$ as follows:
\begin{equation}
    T^{CA} = \mathrm{MLLM}(I,T, p^{CA})
\end{equation}

These four steps constitute a core component of our approach, enabling a multi-level interpretation of memes. Step ID and TM belong to shallow perception, extracting surface information from visual and textual modalities. Step CIM targets deep interpretation, revealing implicit cross-modal meanings. Step CA represents associative reasoning, inferring plausible usage contexts. Together, these four steps significantly enrich the original meme content, empowering downstream classifier to make more accurate predictions.

\subsection{Dual-stage Modal Fusion}

\begin{table*}[htp!]
  \centering
  \renewcommand{\arraystretch}{1.0}
  \resizebox{\textwidth}{!}{
    \begin{tabular}{clcccccccc}
    \toprule
    \multicolumn{1}{c}{\multirow{2}[1]{*}{\textbf{Modality}}} & \multicolumn{1}{l}{\multirow{2}[1]{*}{\textbf{Model}}} & \multicolumn{4}{c}{\textbf{MET-MEME}} & \multicolumn{4}{c}{\textbf{MOOD}} \\
    \cmidrule(lr){3-6} \cmidrule(lr){7-10}
     &  & \textbf{Accuracy} & \textbf{Precision} & \textbf{Recall} & \textbf{Macro-F1} & \textbf{Accuracy} & \textbf{Precision} & \textbf{Recall} & \textbf{Macro-F1} \\
    \midrule
    \multicolumn{1}{c}{\multirow{3}[1]{*}{Unimodal Methods}} & \multicolumn{1}{l}{ResNet-50} & 29.12 & 22.87 & 22.27 & 21.86 & 68.23 & 70.63 & 66.74 & 68.35 \\
    \multicolumn{1}{c}{} & \multicolumn{1}{l}{ViT} & 32.65 & 28.57 & 26.07 & 26.39 & 69.85 & 74.12 & 67.78 & 70.32 \\
    \multicolumn{1}{c}{} & \multicolumn{1}{l}{BERT} & 29.67 & 26.85 & 25.45 & 25.56 & 68.56 & 67.71 & 63.95 & 65.27 \\
    \midrule
    \multicolumn{1}{c}{\multirow{4}[0]{*}{MLLMs}} & \multicolumn{1}{l}{Qwen2.5-VL-7B (zero-shot)} & 34.79 & 40.18 & 35.97 & 32.30 & 56.41 & 60.38 & 51.48 & 48.04 \\
    \multicolumn{1}{c}{} & \multicolumn{1}{l}{Qwen2.5-VL-7B (sft)} & \underline{45.40}  & 41.62 & \underline{41.43} & \underline{41.03} & 62.93 & 65.96 & 54.36 & 54.33 \\
    \multicolumn{1}{c}{} & \multicolumn{1}{l}{Qwen2.5-VL-32B (zero-shot)} & 38.58 & 44.29 & 40.37 & 35.30 & 59.10 & 59.07 & 56.58 & 53.04 \\
    \multicolumn{1}{c}{} & \multicolumn{1}{l}{GPT-4.1 (zero-shot)} & 41.61 & \underline{44.66} & 36.92 & 33.08 & 72.13 & 69.09 & 70.42 & 67.13 \\
    \midrule
    \multicolumn{1}{c}{\multirow{7}[1]{*}{Multimodal Methods}} & \multicolumn{1}{l}{MMBT} & 26.18 & 23.50 & 22.74 & 22.32 & 68.00 & 71.27 & 66.05 & 68.00 \\
    \multicolumn{1}{c}{} & \multicolumn{1}{l}{VisualBERT} & 28.92 & 25.51 & 24.95 & 24.92 & 67.25 & 79.61 & 67.25 & 70.02 \\
    \multicolumn{1}{c}{} & \multicolumn{1}{l}{MET\_add} & 38.72 & 36.23 & 34.96 & 35.33 & 68.19 & 69.73 & 65.17 & 66.93 \\
    \multicolumn{1}{c}{} & \multicolumn{1}{l}{MET\_cat} & 40.07 & 36.37 & 35.79 & 36.01 & 70.32 & 71.83 & 67.76 & 69.51 \\
    \multicolumn{1}{c}{} & \multicolumn{1}{l}{Early Fusion} & 44.27 & 41.04 & 39.76 & 39.94 & \underline{79.48} & 81.48 & 77.11 & 78.93 \\
    \multicolumn{1}{c}{} & \multicolumn{1}{l}{Late Fusion} & 44.05 & 40.01 & 39.94 & 39.61 & 79.38 & \underline{81.77} & 77.23 & 79.02 \\
    \multicolumn{1}{c}{} & \multicolumn{1}{l}{ALFRED} & 34.70 & 34.94 & 34.70 & 34.44 & 79.10 & 81.63 & \underline{79.22} & \underline{79.25} \\
    \midrule
    \midrule
    \multicolumn{1}{c}{} & \multicolumn{1}{l}{\textit{MemoDetector} (Ours)} & \textbf{49.57} & \textbf{47.18} & \textbf{44.91} & \textbf{45.33} & \textbf{83.52} & \textbf{84.59} & \textbf{81.19} & \textbf{82.65} \\
    \bottomrule
    \end{tabular}}
    \caption{Meme emotion detection results on MET-MEME and MOOD datasets. The best and second test results are in bold and underlined, respectively.}
  \label{tab:main}
  \vspace{-4mm}
\end{table*}

\textbf{Feature Extraction.} After obtaining the enhanced textual modality, we first extract features from both image and text. Following the setup of \citet{lin2024towards}, we employ ViT \cite{dosovitskiy2020image} to encode visual features. For text modality, to ensure our model is capable of handling memes in multiple languages, we adopt XLM-R \cite{conneau2019unsupervised} as the text encoder. The overall feature extraction process can be formulated as follows:
\setlength{\abovedisplayskip}{4pt}
\setlength{\belowdisplayskip}{4pt}
\begin{align}
    \mathbf{H_v} &= \mathrm{ViT}(I) \label{eq:vit} \\
    \mathbf{H_\tau} &= \mathrm{XLM\text{-}R}(T), \quad 
    \mathbf{H_\tau^{*}} = \mathrm{XLM\text{-}R}(T^{*}) \label{eq:text}
\end{align}
$I$ and $T$ denote meme image and text. $T^{*}$ denotes the enhanced text. The image is encoded into a sequence of visual patches $\mathbf{H_v} = [v_1, v_2, \ldots, v_n]$, $v_i \in \mathbb{R}^d$, and the original and enhanced texts are encoded into token sequences $\mathbf{H_\tau} = [\tau_1, \tau_2, \ldots, \tau_m]$, $\tau_i \in \mathbb{R}^d$ and $\mathbf{H_\tau^{*}} = [\tau^*_1, \tau^*_2, \ldots, \tau^*_{m^*}]$, $\tau_i^* \in \mathbb{R}^d$, respectively. Here, $* \in \{\text{ID}, \text{TM}, \text{CIM}, \text{CA}\}$ corresponds to the four enhancement steps.

\noindent\textbf{Modal Fusion.} The interplay between modalities in memes is often subtle and implicit, making it difficult for traditional single-stage fusion methods to achieve fine-grained alignment. This may lead to misinterpretation of emotional signals. To address this, we propose a dual-stage fusion strategy. In the first stage, we perform a shallow fusion of surface-level features from both modalities, enabling the small model to form an initial, coarse understanding. In the second stage, we deepen this understanding by integrating the shallow features with the enriched, multi-level semantic cues extracted by MLLM. This hierarchical design enables the model to progressively enhance its interpretation of cross-modal signals. We now detail the fusion process.

\textbf{Stage 1.} The primary goal of this stage is to achieve an initial alignment between the shallow features of the visual and textual modalities. Specifically, we concatenate the extracted image patch sequence $\mathbf{H_v}$ with the original text token sequence $\mathbf{H_\tau}$, resulting in an enhanced image patch sequence $\mathbf{H_v^{'}}$ that incorporates information from both modalities.
\begin{equation}
    \mathbf{H_v^{'}} = \mathrm{Concat}(\mathbf{H_v},\mathbf{H_\tau}) = [v_1, \ldots, v_n, \tau_1, \ldots, \tau_m]
\end{equation}
The enhanced visual representation $\mathbf{H_v^{'}}\in \mathbb{R}^{N \times d}$ consists of $n$ original image patches and $m$ text tokens. Here, the text tokens can be viewed as special ``pseudo-patches" that highlight the textual modality and integrate it into the visual representation. By incorporating these tokens, the fused feature $\mathbf{H_v^{'}}$ embeds initial multimodal information into a unified representation.

\textbf{Stage 2.} After obtaining $\mathbf{H_v^{'}}$, we proceed with the second-stage fusion by integrating it with the enriched textual tokens derived from MLLM. First, we concatenate the tokens generated from the four-stage enhancement process to form a unified enriched textual representation $\mathbf{H_\tau^{'}} \in \mathbb{R}^{M \times d}$:
\begin{equation}
    \mathbf{H_\tau^{'}} = \mathrm{Concat}(\mathbf{H_\tau^{\text{ID}}}, \mathbf{H_\tau^{\text{TM}}}, \mathbf{H_\tau^{\text{CIM}}}, \mathbf{H_\tau^{\text{CA}}})
\end{equation}
Next, we employ a bidirectional cross-attention mechanism to deeply fuse the enhanced visual representation $\mathbf{H_v^{'}}$ and enriched textual representation $\mathbf{H_\tau^{'}}$. Specifically, to attend visual representations to textual ones, we compute the attended visual features $\mathbf{H_{v}^{\text{att}}}$ as follows:
\begin{equation}
    \mathbf{H_{v}^{\text{att}}} = \mathrm{softmax}\left (\frac{Q_{\tau}K_{v}^{\top}}{\sqrt{d_k} } \right ) V_{v}
\end{equation}
where $\{Q_{\tau},K_{v},V_{v}\} = \{\mathbf{H_\tau^{'}}W_Q,\mathbf{H_v^{'}}W_K,\mathbf{H_v^{'}}W_V\}$. Then we fuse $\mathbf{H_{v}^{\text{att}}}$ with $\mathbf{H_\tau^{'}}$ to attain cross-modally enhanced textual representation $\boldsymbol{\tilde\tau}$:
\begin{equation}
    \boldsymbol{\tilde\tau} = \mathbf{H_\tau^{'}} + \mathbf{H_{v}^{\text{att}}}W_O
\end{equation}
where $W_O$ denotes linear projection. Similarly, we can attain cross-modally enhanced visual representation $\mathbf{\tilde v}$:
\begin{equation}
    \mathbf{\tilde v} = \mathbf{H_v^{'}} + \text{softmax}\left (\frac{Q_{v}K_{\tau}^{\top}}{\sqrt{d_k} } \right ) V_{\tau}W_{O}^{'}
\end{equation}
After attaining $\mathbf{\tilde v}$ and $\boldsymbol{\tilde\tau}$, we concatenate these two vectors to obtain the final fused representation of the meme:
\begin{equation}
    \mathbf{E^{\text{meme}}} = [\text{mean}(\mathbf{\tilde v}), \text{mean}(\boldsymbol{\tilde\tau})]
\end{equation}
Finally, we use a linear layer and a softmax classifier for meme emotion classification:
\begin{equation}
    \hat{y} = \text{softmax}(W\mathbf{E^{\text{meme}}} + b)
\end{equation}

\noindent\textbf{Training.} The training objective is to minimize the average cross-entropy loss between predictions and ground-truth labels across the dataset. The loss function is as follows:
\begin{equation}
    \mathcal{L} = \frac{1}{\left | \mathcal{D} \right | } \sum_{i=1}^{\left | \mathcal{D} \right |} L_{CE}(\hat{y}, y)
\end{equation}
where $\mathcal{D}$ is the number of samples in the training set, with $\hat{y}$ and $y$ representing predicted label and true label. $L_{CE}$ is the cross-entropy loss function.

\section{Experiments}

\subsection{Experimental Setup}

\textbf{Datasets.} We evaluate on two public meme datasets: MET-MEME \cite{xu2022met}, a bilingual dataset with 4,000 English and 6,045 Chinese memes labeled across seven emotion classes, and MOOD \cite{sharma2024emotion}, which contains 10,004 English memes annotated with six emotions.

\noindent \textbf{Baselines.} We compare our model against several state-of-the-art approaches for meme emotion detection. These baseline models can be broadly categorized into three groups. (1) Unimodal methods that utilize only visual or textual information, such as ResNet \cite{he2016deep}, ViT \cite{dosovitskiy2020image} and BERT \cite{devlin2019bert}. (2) MLLM-based methods. We evaluate open-source models including Qwen2.5-VL-7B and Qwen2.5-VL-32B \cite{bai2025qwen2}, as well as the commercial GPT-4.1. (3) Multimodal methods based on small models. These include MMBT \cite{kiela2019supervised}, VisualBERT \cite{li2019visualbert}, MET\_add, MET\_cat \cite{xu2022met}, Early Fusion \cite{pranesh2020memesem}, Late Fusion \cite{pramanick2021detecting} and ALFRED \cite{sharma2024emotion}.

\noindent \textbf{Evaluation Metrics.} We report accuracy and macro-F1 as primary metrics, along with macro-averaged precision and recall for completeness. All our scores are the average over 5 runnings with random seeds.

\subsection{Main Results}

Table \ref{tab:main} presents the performance of our method against all baselines on MET-MEME and MOOD. Our approach consistently outperforms all SoTA baselines across all metrics on both datasets, highlighting several key insights. (1) Multimodal methods generally outperform unimodal ones, as they better exploit both visual and textual signals for meme understanding. However, the effectiveness of these models largely depends on the fusion strategy. Poorly designed fusion can lead to inferior performance, as seen with MMBT, which underperforms both ViT and BERT on MET-MEME in terms of Macro-F1. (2) Although MLLMs have the largest number of parameters, their zero-shot performances fall short of expectation, likely due to their inherent limitations and unsuitability for direct classification tasks. However, fine-tuning significantly improves results—Qwen2.5-VL-7B even surpasses GPT-4.1 on MET-MEME, suggesting that MLLMs possess strong potential for deep meme understanding when properly adapted. (3) Both Early Fusion and Late Fusion outperform other baselines across the two datasets, with ALFRED demonstrating notably strong performance on the MOOD dataset. These superior performances can be attributed to the fact that the encoders for both modalities are well pre-trained, enabling them to effectively capture emotion-related features. When combined with relatively effective fusion strategies, these models are able to produce more accurate emotion predictions. (4) Notably, our model significantly surpasses these baseline systems. On MET-MEME, it surpasses the Early Fusion method by 5.3\%, 6.14\%, 5.15\%, and 5.39\% in terms of accuracy, precision, recall, and macro-F1, respectively. On MOOD dataset, it achieves improvements of 4.42\% in accuracy, 2.96\% in precision, 1.97\% in recall, and 3.4\% in macro-F1 compared to ALFRED. These findings suggest that by incorporating MLLM-generated multi-level meme analysis for textual enrichment, combined with carefully designed dual-stage modal fusion strategy, we achieve substantial performance gains in meme emotion detection.

\begin{figure*}[t]
\centering
\includegraphics[width=0.8\textwidth]{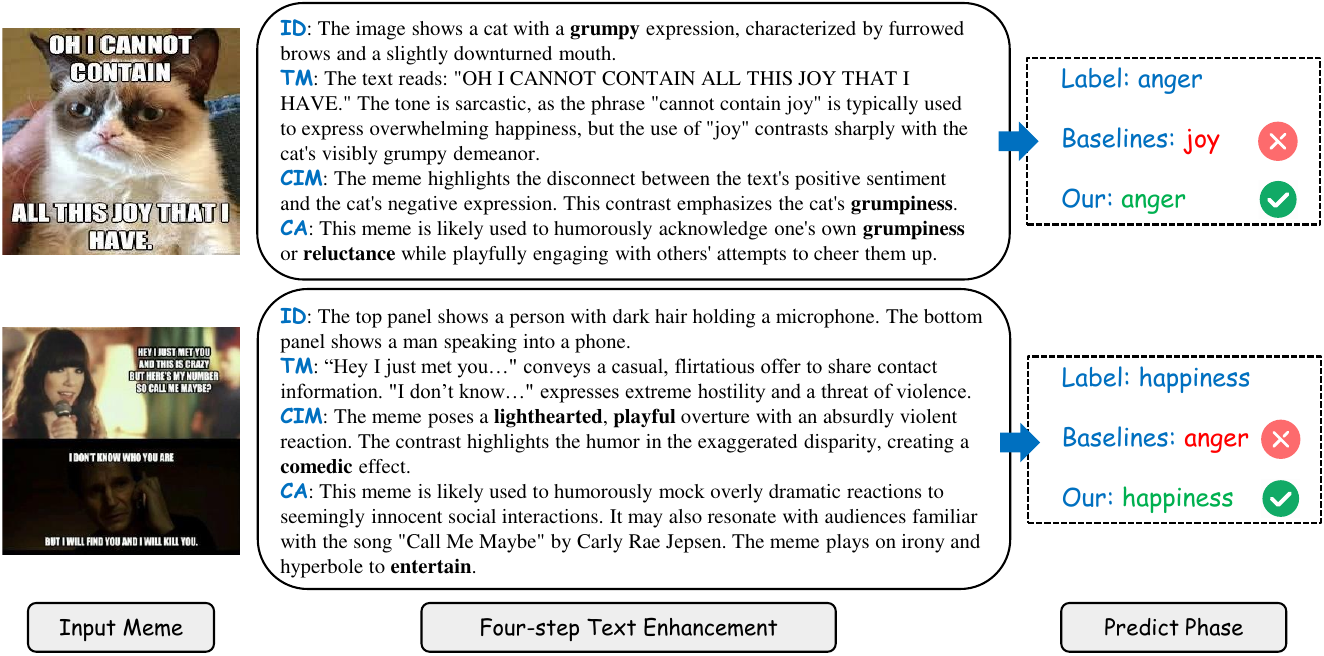} 
\caption{Examples of correctly predicted memes.}
\label{case}
\vspace{-4mm}
\end{figure*}

\begin{table}
  \centering
  \resizebox{0.475\textwidth}{!}{
    \begin{tabular}{c|lcccc}
    \toprule
    \textbf{Dataset} & \textbf{Method} & \textbf{Accuracy} & \textbf{Precision} & \textbf{Recall} & \textbf{Macro-F1} \\
    \midrule
    \multirow{6}[1]{*}{\textbf{MET-MEME}} & Ours  & \textbf{49.57} & \textbf{47.18} & \textbf{44.91} & \textbf{45.33} \\
    \cdashline{2-6}
          & w/o DF & 48.57 & 46.51 & 43.35 & 43.57 \\
          & w/o ID & 48.84 & 45.61 & 44.23 & 44.29 \\
          & w/o TM & 47.82 & 46.35 & 43.06 & 43.78 \\
          & w/o CIM & 48.63 & 46.08 & 43.91 & 44.32 \\
          & w/o CA & 48.22 & 45.16 & 43.82 & 43.97 \\
    \midrule
    \multirow{6}[1]{*}{\textbf{MOOD}} & Ours  & \textbf{83.52} & \textbf{84.59} & \textbf{81.19} & \textbf{82.65} \\
    \cdashline{2-6}
          & w/o DF & 82.32 & 82.93 & 80.23 & 81.39 \\
          & w/o ID & 82.30 & 82.29 & 80.56 & 81.32 \\
          & w/o TM & 81.62 & 81.68 & 79.72 & 80.54 \\
          & w/o CIM & 81.73 & 82.95 & 79.25 & 80.75 \\
          & w/o CA & 81.81 & 81.85 & 79.87 & 80.75 \\
    \bottomrule
    \end{tabular}}
    \caption{Ablation studies by removing components from our proposed framework. "DF" denotes Dual-stage Fusion module, while "ID", "TM", "CIM", and "CA" correspond to the four steps in textual modality enhancement, respectively.}
  \label{tab:ablation}
  \vspace{-4mm}
\end{table}

\subsection{Ablation Study}

We perform ablation studies to evaluate the contribution of each component in our model. As shown in Table \ref{tab:ablation}, our full model, without any component removed, achieves the best performance across all metrics, clearly demonstrating the effectiveness of each proposed module. Specifically, when the dual-stage fusion strategy is ablated—replacing it with a simple concatenation of the enhanced textual representation and visual features—the model's performance drops significantly across both datasets (with a notable 1.76\% decrease in F1 
on MET-MEME). This highlights the importance of our dual-stage fusion design, which captures the intricate interactions between visual and textual modalities in memes and enables the model to progressively develop a deeper emotional understanding. Moreover, removing any of the four textual enhancement steps leads to a performance drop to varying degrees, which confirms the contribution of each step. Both shallow-level enhancements such as ID and TM, and deeper-level reasoning components like CIM and CA, provide valuable information that boosts the model’s emotion detection performance. Notably, removing the TM module results in the most significant performance degradation (a 1.55\% drop in F1 on MET-MEME and 2.11\% on MOOD), highlighting the crucial role of understanding meme text in capturing its emotional implications.

\subsection{Case Study}

To more clearly demonstrate the effectiveness of our method, we present two representative examples, as shown in Figure \ref{case}. The first meme presents a sarcastic contrast between the text and the image of an obviously angry cat, which collectively conveys a sense of displeasure rather than genuine joy. Baseline methods, lacking a fine-grained modal fusion mechanism, are misled by the overtly positive textual sentiment and thus fail to predict the correct emotion. In contrast, our method effectively integrates both the surface content and the underlying multimodal cues through a carefully designed dual-stage fusion strategy, enabling it to correctly identify the intended emotional tone. The second meme contrasts the lighthearted lyrics of an English song with an exaggerated overreaction from a person, thereby creating a humorous and playful atmosphere. Baseline methods, lacking knowledge about the song lyrics, fail to capture the intended meaning. In contrast, our method leverages a four-stage text enhancement process, where MLLM contributes rich contextual understanding to uncover the humorous intent behind the lyrics. This enhanced textual representation effectively guides the small model to make the right prediction.

\subsection{Analyses on Proposed Methods}

To gain deeper insights into the effectiveness of our proposed approach, we perform a series of in-depth analyses aimed at addressing the following three key questions.

\begin{figure}[t]
\centering
\includegraphics[width=0.49\textwidth]{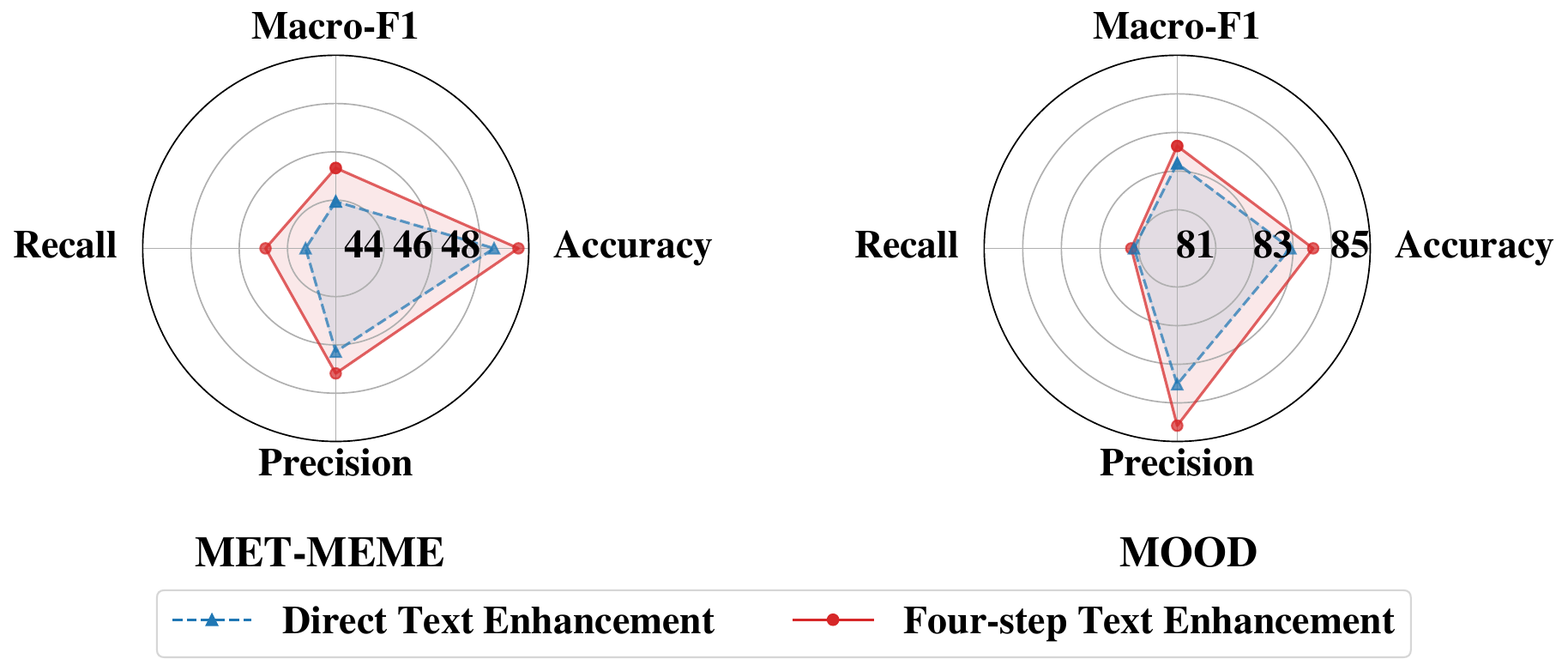} 
\caption{Comparison between four-step text enhancement strategy and direct text enhancement strategy.}
\label{diff_enhance}
\end{figure}

\noindent \textbf{Q1: What are the advantages of four-step text enhancement?} In Table~\ref{tab:ablation}, we have already shown that each step of the proposed four-step text enhancement contributes positively to the overall performance. To further investigate its true advantage, we introduce an alternative approach called direct textual enhancement for comparison. In this setting, MLLM is prompted to directly infer the meme’s emotional tendency and provide a single-step explanation, which is then used as the enhanced text. As shown in Figure~\ref{diff_enhance}, our four-step enhancement strategy consistently outperforms the direct method across all metrics on both datasets. This suggests that our structured, progressive prompting enables the MLLM to develop a deeper and more comprehensive understanding of the meme. The resulting interpretive texts are of higher quality and thus more effective in guiding the small model to make accurate predictions.

\begin{table}[t]
  \centering
  \resizebox{0.45\textwidth}{!}{
    \begin{tabular}{l||cc|cc}
    \toprule
    \multicolumn{1}{l}{\textbf{Dataset}} & \multicolumn{2}{c}{\textbf{MET-MEME}} & \multicolumn{2}{c}{\textbf{MOOD}} \\
    \midrule
    \textbf{Model} & \textbf{Accuracy} & \textbf{Macro-F1} & \textbf{Accuracy} & \textbf{Macro-F1} \\
    \midrule \midrule
    QwenVL7B & 34.79 & 32.30 & 56.41 & 48.04 \\
    + CoT & 40.77 & 36.59 & 61.05 & 55.41 \\
    + Ours & 45.60 & 41.22 & 82.70 & 81.72 \\
    \midrule
    QwenVL32B & 38.58 & 35.30 & 59.10 & 53.04 \\
    + CoT & 38.18 & 32.74 & 55.00 & 49.31 \\
    + Ours & 49.57 & 45.33 & 83.52 & 82.65 \\
    \bottomrule
    \end{tabular}}
    \caption{Effect of MLLM scale and usage paradigm on meme emotion detection performance.}
  \label{tab:diff_model}
  \vspace{-4mm}
\end{table}

\noindent \textbf{Q2: How do the scale of MLLMs and usage paradigms influence the performance of meme emotion detection?} In our framework, QwenVL-32B is adopted by default to perform text enhancement for memes. To investigate how the scale of MLLM and the usage paradigm affect the final performance, we additionally introduce a smaller model, QwenVL-7B, and design three distinct usage paradigms for comparison, as presented in Table \ref{tab:diff_model}. Specifically, we explore three paradigms: (1) QwenVL7B/32B: Prompt MLLM directly to generate the emotion label for a given meme. (2) + CoT: Incorporate chain-of-thought prompting to encourage step-by-step reasoning before producing the final answer. (3) + Ours: Apply our proposed framework, where MLLM performs multi-level textual augmentation of memes to help a smaller model in emotion prediction. We can observe that: (1) The direct deployment of both MLLMs struggles since large models are not specifically designed for this classification task \cite{xu2024llms,li2024task}. (2) CoT prompting significantly improves the performance of QwenVL7B, whereas QwenVL32B exhibits a performance drop. A possible explanation is that smaller models benefit from explicit reasoning structures due to their limited capacity for complex multimodal inference. In contrast, larger models may already possess sufficient reasoning capabilities, and step-by-step constraints might inhibit their natural inference process, leading to degraded performance. (3) Our approach focuses on training a small model, which not only avoids the impractical cost of fine-tuning large models but also benefits from the insights provided by them, ultimately leading to improved performance. (4) Our method achieves better results when using the 32B model for textual enhancement compared to the 7B variant. This suggests that larger models may generate higher-quality interpretative text, which in turn provides more effective guidance for downstream classification.

\begin{figure}[t]
\centering
\includegraphics[width=0.475\textwidth]{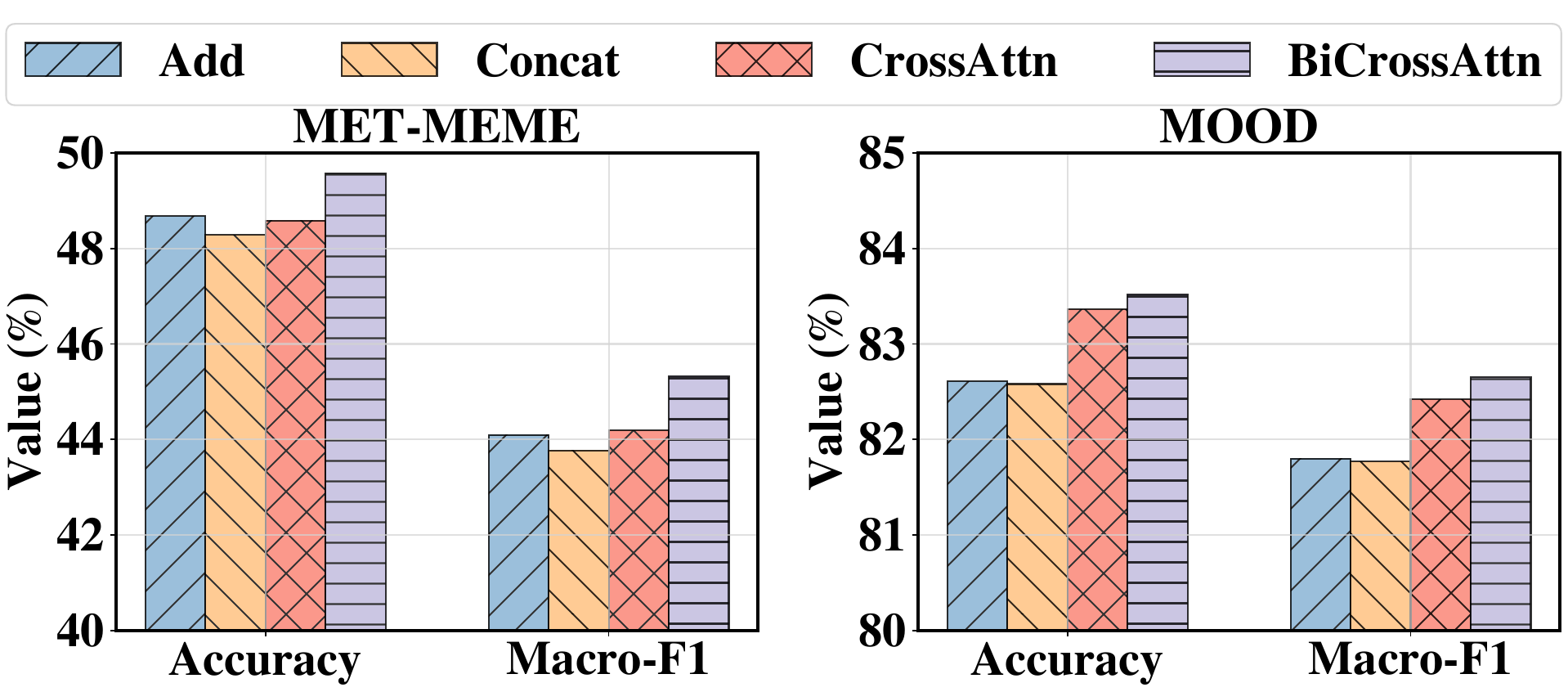} 
\caption{Impact of different second-stage fusion strategies on model performance.}
\label{diff_fusion}
\vspace{-4mm}
\end{figure}

\noindent \textbf{Q3: Why do we design a bidirectional cross-attention mechanism in the second stage of modal fusion?} In ablation study, we have demonstrated that our dual-stage fusion strategy outperforms conventional single-stage fusion approaches. Specifically, in the second fusion stage, we employ a bidirectional cross-attention mechanism to integrate the enhanced visual and textual features. To further investigate the superiority of this design, we replace the bidirectional cross-attention module with several commonly used modal fusion strategies (i.e., add, concatenate, and cross-attention). The comparison results are shown in Figure~\ref{diff_fusion}. We can observe that our proposed bidirectional cross-attention consistently outperforms other fusion strategies across both datasets, demonstrating the effectiveness and superiority of our design. This mechanism enables the enhanced textual features to attend to the enriched visual representations and vice versa, facilitating a more comprehensive and interactive integration of multimodal information. As a result, the model can better capture the intricate relationships between modalities, ultimately leading to improved performance.

\section{Conclusion}

In this paper, we address two core challenges in MEU: the lack of fine-grained multimodal fusion strategies and insufficient mining of memes’ implicit meanings and background knowledge. To tackle these issues, we propose \textit{MemoDetector}, a novel framework that leverages MLLMs to deeply interpret meme semantics and contextual cues. In addition, we design a dual-stage modal fusion strategy that performs both shallow and deep integration of visual and textual features, enabling more nuanced emotional understanding. Extensive experiments and thorough analyses demonstrate the effectiveness, robustness, and SoTA performance of our method.

\section{Acknowledgments}
This research received support from the National Natural Science Foundation of China under Grant No. 62302441. This work was also supported by the Key Research and Development Program Project of Ningbo Grant No. 2025Z029. The author gratefully acknowledges the support of Zhejiang University Education Foundation Qizhen Scholar Foundation.

\bibliography{aaai2026}

\end{document}